\documentclass[runningheads]{llncs}

\usepackage{eccv}

\usepackage{eccvabbrv}

\usepackage{graphicx}
\usepackage{booktabs}
\usepackage{booktabs}
\usepackage[accsupp]{axessibility}  % Improves PDF readability for those with disabilities.
\usepackage{booktabs}
\usepackage{colortbl}
\usepackage{xcolor}
\usepackage{tabularx}
\definecolor{brightyellow}{rgb}{1,1,0.6}
\definecolor{brightcyan}{rgb}{0.6,1,1}

\usepackage{xcolor}
\newcommand{\papername}{{\color{black}SweepNet }} % Define \papername with blue color

\newcommand{\mypara}[1]{\noindent \textbf{\textit{#1}.}}
\usepackage{hyperref}

\usepackage{orcidlink}
\usepackage{wrapfig}
\usepackage{array}
\begin{document}

\title{SweepNet: Unsupervised Learning Shape Abstraction via Neural Sweepers} 

\titlerunning{SweepNet}

\author{Mingrui Zhao\inst{1} \and
Yizhi Wang\inst{1} \and
Fenggen Yu\inst{1} \and
Changqing Zou \inst{2} \and \\
Ali Mahdavi-Amiri \inst{1}}

\authorrunning{M.~Zhao et al.}

\institute{Simon Fraser University \and
Zhejiang University\\}
% \email{mza143@sfu.ca}

%\maketitle

{
\maketitle
\vspace{2em}
\centering
\includegraphics[width=1\textwidth]{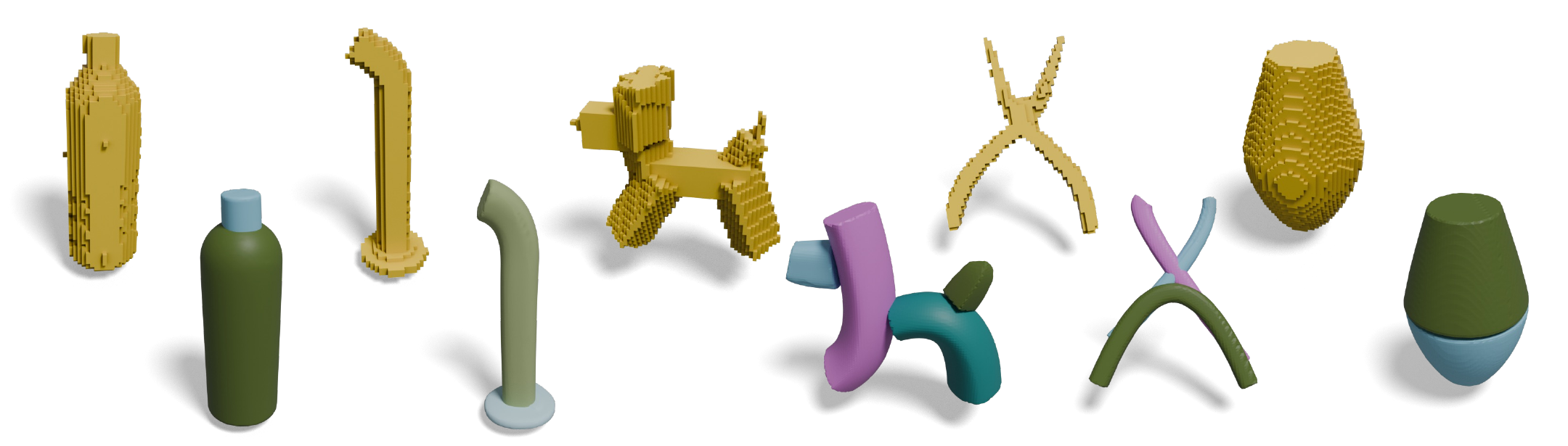}
\captionof{figure}{
Given a voxelized shape, \papername acquires abstraction without any supervision. With few sweep surfaces, the overall geometry of the objects is nicely captured.
}
\vspace{1em}
\label{fig:teaser}
}

\begin{abstract}

Shape abstraction is an important task for simplifying complex geometric structures while retaining essential features. Sweep surfaces, commonly found in human-made objects, aid in this process by effectively capturing and representing object geometry, thereby facilitating abstraction. In this paper, we introduce \papername, a novel approach to shape abstraction through sweep surfaces. We propose an effective parameterization for sweep surfaces, utilizing superellipses for profile representation and B-spline curves for the axis. This compact representation, requiring as few as 14 float numbers, facilitates intuitive and interactive editing while preserving shape details effectively. Additionally, by introducing a differentiable neural sweeper and an encoder-decoder architecture, we demonstrate the ability to predict sweep surface representations without supervision. We show the superiority of our model through several quantitative and qualitative experiments throughout the paper. Our code is available at \url{https://mingrui-zhao.github.io/SweepNet/}.

\end{abstract}

\section{Introduction}
\label{sec:intro}

Sweep surfaces play an important role in computer graphics and computer vision, serving as fundamental constructs for modelling and analyzing complex shapes and structures. Sweep surfaces are extensively utilized for generating intricate geometric forms by sweeping a cross-sectional profile along a defined path. This enables the creation of diverse objects ranging from simple curves to intricate architectural designs, and more. 
On the other hand, shape abstraction is also an important problem in computer vision and graphics that involves representing complex geometric structures or objects in a simplified form while preserving essential characteristics for analysis or visualization purposes. Sweep surfaces can serve as a powerful tool for shape abstraction due to their ability to efficiently capture and represent the geometry of objects or structures due to their generality and ubiquity in the objects existing around us. By utilizing sweep surfaces, 3D shapes can be abstracted into more manageable representations, facilitating tasks such as shape recognition, and manipulation in various domains including computer graphics, or computer-aided design. 

Current approaches to shape abstractions can be categorized by the type of constituent primitives. Popular choices include cuboid \cite{tulsiani2017learning, yang2021unsupervised, smirnov2020deep, zou20173d}, superquadrics \cite{paschalidou2019superquadrics, yu2022capri, liu2023marching, wu2022primitive, sharma2022prifit, liu2022robust}, parametric surfaces \cite{sharma2020parsenet, yang2023neural}, convex shapes \cite{chen2020bsp, deng2020cvxnet}, neural parts \cite{paschalidou2021neural, kawana2020neural, huang2023learning}, sketch-and-extrude \cite{li2023secad, ren2022extrudenet, uy2022point2cyl}, and a combination of simple primitives \cite{li2019supervised, hao2020dualsdf, sharma2018csgnet, ren2021csg, kania2020ucsg, li2023surface}.  However, each representation undergoes unique advantages and limitations. For instance, parametric primitives like cuboids and superquadrics offer ease of modification through parameter adjustments, facilitating interactivity. However, their simplicity often leads to less compact and expressive abstractions. Conversely, neural primitives showcase superior expressiveness by capturing complex shapes more accurately but suffer from reduced manipulability post-creation, which limits user control.

Shape abstraction via sweep surfaces can strike a balance between compactness and preserving details (see Fig.~\ref{fig:teaser}). 
However, learning shape abstraction via sweep surfaces is challenging, primarily due to the limitations of accurate representation and the complexities involved in their parametrization. Existing methods of determining sweep surfaces are formulated as optimization problems \cite{sellan2021swept} concerning sweep profile, sweep axis, and sweep motion, which makes it hard to integrate within a larger problem or a deep neural network. When integrated into computational methods, this introduces a nested optimization challenge, and within a learning framework, it results in a non-differentiable component.

In this paper, we first provide a simple parameterization for sweep surfaces that can be learned via a differentiable network. We employ superellipses for the profile representation due to its simplicity to learn and diversity in shape (see Fig.~\ref{fig: superellipse}). For the axis, we use B-spline curves, in conjunction with basic polynomials to control sweep dynamics. Consequently, in our representation, a sweep surface can be represented with as few as 14 float numbers. Moreover, the complexity of these primitives can be easily scaled by expanding the parameter space. We then show how we can utilize this representation to learn shape abstractions for a given 3D object. Our approach involves an encoder-decoder architecture and a differentiable neural sweeper, enabling the model to predict the sweep surface representations \textit{without any supervision}. By jointly optimizing representation faithfulness, sweeping rationality, and primitive parsimony, our model delivers high-quality abstractions of the target shape (Fig.~\ref{fig: pipeline} and Fig.~\ref{fig:teaser}). Therefore, our contributions are as follows:
\begin{itemize}
    \item We provide the first deep learning model equipped with a differentiable sweeper specifically designed for shape abstraction through sweep surfaces.
    \item Our method introduces a new and compact parameterization of sweep surfaces, enabling intuitive and interactive editing.
    \item We demonstrate the advantages of our sweep surfaces over traditional parametric primitives in representing curvy-featuring objects, showcasing its superiority in achieving concise and expressive shape abstractions.
\end{itemize}

\section{Related Work}
\label{sec:related works}
\mypara{Swept Volumes}
Swept volume~\cite{sungurtekin1986graphical,wang1986geometric,abdel2006swept,rossignac2007boundary,zhang2009reliable,campen2010polygonal,sellan2021swept} refers to the total volume displaced by a moving object as it travels through a particular path or trajectory.
The key challenges of constructing swept volumes involve not only constructing ruled surface patches for each edge and face but also trimming their mutual intersections to remove components not contributing to the final surface.
S\`ellan et al.~\cite{sellan2021swept} introduce spacetime numerical continuation for swept volume construction, offering enhanced generality and robustness with asymptotic complexity one order lower than prevailing industry standards.
However, the construction process of swept volumes is typically non-differentiable, precluding its integration into our network. To address this challenge, we propose Neural Sweeper, a neural network designed to approximate implicit fields for swept volumes using profile and axis information as input. We leverage the methodology outlined in~\cite{sellan2021swept} for data preparation, wherein ground-truth mesh data of swept volumes is generated to calculate the occupancy field for training our neural network.

\mypara{Neural Implicit Fields}
OccNet~\cite{mescheder2019occupancy}, IM-Net~\cite{IM-Net}, and DeepSDF~\cite{park2019deepsdf} concurrently 
introduced the coordinate-based neural implicit representation. 
These early works only model global shape features, yielding over-smooth results which lack geometric details.
The next wave in this direction has focused on conditioning implicit neural representations on local features stored in 
voxel~\cite{Chibane2020,Jiang2020_LocalGrid,Peng2020_CON,Chabra2020_DeepLS,tang2021sa}, image grids~\cite{PIFu,DISN,D2IM-Net} or surface points~\cite{boulch2022poco,giebenhain2021air,williams2022neural,huang2023neural}
to more effectively recover geometric or topological details and to scale to scene reconstruction, % the representation learning from the object level to scene level, 
or at the patch level~\cite{LDIF,PatchNets,zhang20223dilg,wang2023aro} to improve generalizability across object categories. 
We utilize POCO~\cite{boulch2022poco} as the backbone network of our neural sweeper, which takes as input the point cloud of the sweep surface to predict its implicit field.

\mypara{Primitive Detection and Fitting}
Traditional methods~\cite{li2011globfit,schnabel2007efficient,borrmann20113d,rabbani2005efficient,oesau2016planar,drost2015local} for primitive detection involve RANSAC~\cite{fischler1981random} and Hough Transform~\cite{hough1959machine}.
The work of Zou et al.~\cite{zou20173d} and Tulsiani et al~\cite{tulsiani2017learning} are among the earliest works that employ neural networks for primitive fitting, using cuboids as the only primitives.
SPFN~\cite{li2019supervised} and ParSeNet~\cite{sharma2020parsenet} consider unions of primitive patches to fit given 3D objects typically represented by point clouds.
Constructive solid geometry (CSG) is a classical CAD representation which models a 3D shape as a recursive assembly of solid primitives using operations including union, intersection, and difference.
Many recent works of primitive fitting are built on CSG trees, such as
CSGNet~\cite{sharma2018csgnet},
UCSG-NET~\cite{kania2020ucsg},
BSP-Net~\cite{chen2020bsp},
CSG-Stump~\cite{ren2021csg},
CAPRI-Net~\cite{yu2022capri},
and D2CSG~\cite{yu2024d}.
Inspired by the user-level construction sequence of CAD models~\cite{wu2021deepcad},
Point2Cyl~\cite{uy2022point2cyl} and SECAD-Net~\cite{li2023secad} utilize sketch-and-extrude operations which enable the construction of 3D solid shapes from 2D sketches, which can ease the process of primitive fitting.
We introduce a customized sketch-and-extrude process to create our sweep surfaces, with a superellipse as the 2D profile, a B-spline \textbf{curve} as the sweeping axis (direction) and a scaling function re-scaling the profile along the sweeping axis.

\mypara{Shape Abstraction}
Shape abstraction aims to fit 3D objects using \textit{simple} and \textit{compact} geometric primitives.
Computational approaches~\cite{barr1981superquadrics,pentland1987perceptual,gross1988error,solina1990recovery,leonardis1997superquadrics,chevalier2003segmentation,vaskevicius2017revisiting,liu2022robust,wu2022primitive,liu2023marching} directly optimize the parameters of primitives to fit a given shape. The primitives are typically superquadrics due to its extensive shape vocabulary including
cuboids, ellipsoids, cylinders, octohedra, and many
shapes in between.
However, computational approaches typically rely on dense inputs (point clouds or SDFs) and require a closed-form equation for the implicit function of the primitives.
Learning-based approaches~\cite{tulsiani2017learning, sun2019learning, yang2021unsupervised, hao2020dualsdf, sharma2022prifit, paschalidou2020learning, paschalidou2019superquadrics,genova2019learning,vora2024divinet} are versatile
in dealing with different input sources, such as point clouds, voxel grids, or even RGB images.
Since the implicit function of sweep surfaces has no closed-form equations, we learn a neural implicit field for them to approximate the function.

\section{Method}
\label{sec: method}
\begin{figure}[t]
    \centering
    \includegraphics[width=\linewidth]{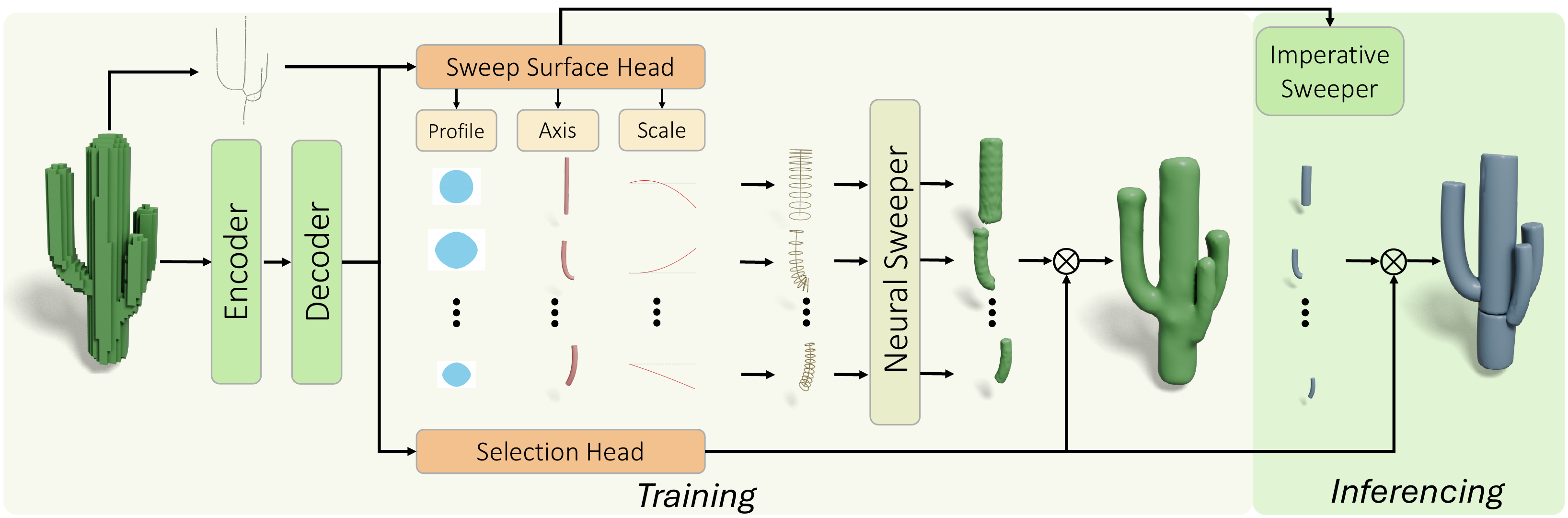}
    \caption{\textbf{Pipeline overview}. The model processes voxel input to extract a skeletal prior and encodes the data with a voxel encoder. The sweep surface head predicts sweep surface parameters: 2D profiles, 3D sweeping axes, and profile scaling function coefficients, conditioned on the skeletal prior. Training involves generating point clouds for each sweep surface through a differentiable sampler, which the neural sweeper uses to estimate their occupancy. This data is then assembled to reconstruct the input shape to quantify loss. At inference time, the sweep surface parameters are directly processed by a non-differentiable, imperative sweeper to produce the resembled shape.}
    \label{fig: pipeline}
\end{figure}

This section presents an unsupervised model that parses 3D shapes using parametric sweep surfaces. We call our model \papername whose overall pipeline is outlined in \cref{fig: pipeline}. It encodes the input voxel shape using a 3D convolutional network to extract features, which are then enhanced through a three-layer MLP to produce a latent representation, $z$. This representation feeds into a dual MLP head: one predicting parameters for multiple sweep surfaces (sweep surface head) and another selecting a subset of these primitives (selection head) for parsimonious shape assembly. During this process, the prediction of the sweeping axes is guided by the medial axis~\cite{tagliasacchi2012mean} of the input voxel shape. For each sweep surface, we select a series of points along its sweeping axis as well as a number of profile slices for examination. These profile slices are created by projecting the 2D profile onto the 3D sweeping curve, utilizing the curve's coordinate frame for transformation. The resulting point cloud is a compilation of points from both the sweeping axis and these transformed loops. This gathered data is then processed by the neural sweeper to predict the occupancy field of the sweep surface. The final assembled shape is presented by the union of the selected sweep surface occupancies. At inference time, the predicted primitive parameters are used directly to produce the final sweep surfaces using standard but non-differentiable sweepers, bypassing the neural sweeper, for efficient primitive production, and the parsed shape is compactly described by their parameters in the size of only tens of floats.

\begin{figure}[t]
    \centering
    \includegraphics[width=\linewidth]{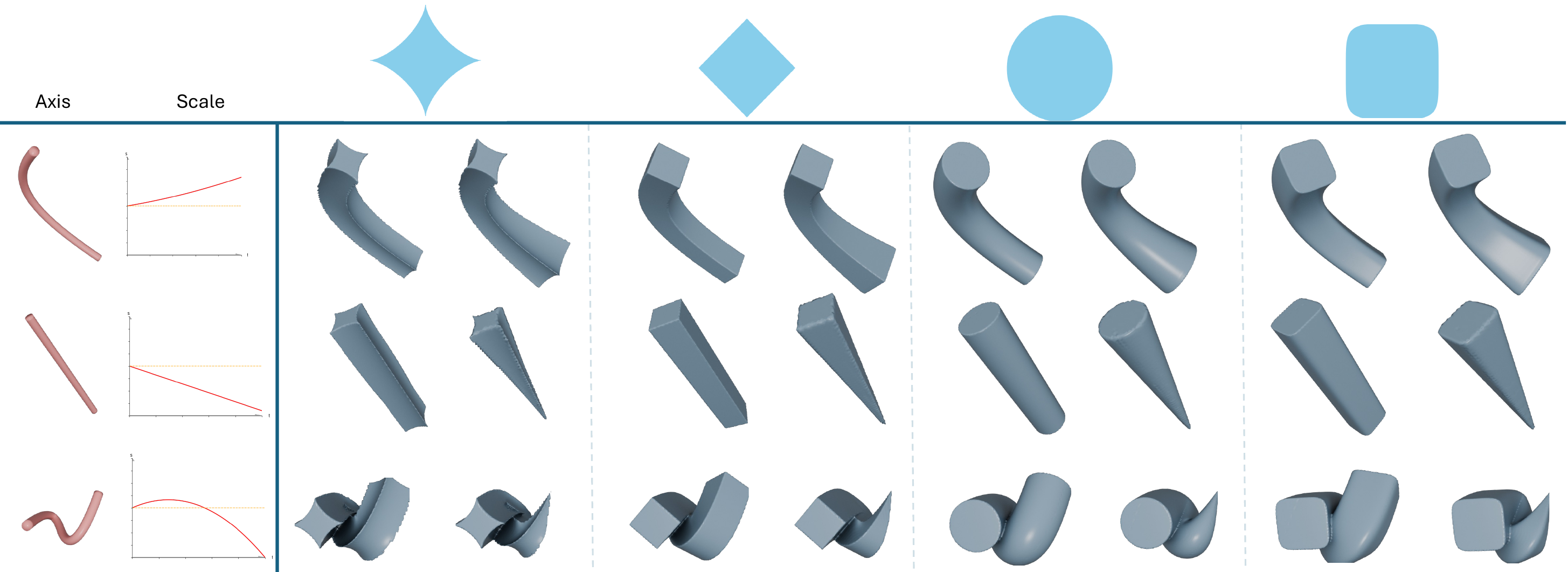}
    \caption{Sweep surface primitives parameterized with different profiles, axis, and scaling functions. The 2D profiles (superellipses) are shown in blue. Constant and dynamic scales of sweep surfaces are shown in alternating columns with respect to each profile and axis pair.}
    \label{fig: superellipse}
    \vspace{-10pt}
\end{figure}

\subsection{Sweep Surface Primitive}
A sweep surface is defined by a 2D profile, a 3D sweeping axis, and a function $f$ controlling the profile's scale along the axis. Here, we target for a parameterizable, compact and expressive representation for all three components.  
A 2D profile is defined as a finite closed loop with no self-intersections. Existing works attempted implicit fields \cite{li2023secad}, neural sketches~\cite{uy2022point2cyl} and rational Bézier polygons \cite{ren2022extrudenet} to represent such profiles. Implicit fields can well represent complex profiles, however, they are carried through a neural network and hence offer limited editability post-creation. The rational Bezier polygon is parameterizable and offers good expressiveness. However, it needs additional constraints in formulation to maintain the non-self-intersection property, requires critical coefficients, and is sometimes impossible, to represent certain regular shapes such as rectangles. Alternatively, we choose superellipse to parameterize the 2D profile, formulated as
\begin{equation}
\left\{
\begin{aligned}
x(\theta) &= a \cdot |\cos(\theta)|^{\frac{2}{d}} \cdot \mathrm{sgn}(\cos(\theta)), \\
y(\theta) &= b \cdot |\sin(\theta)|^{\frac{2}{d}} \cdot \mathrm{sgn}(\sin(\theta)),
\end{aligned}
\right.
\end{equation}
where $\theta \in [0, 2\pi]$ is the polar angle, $x,y \in \mathbb{R}$ are the Cartesian superellipse contour coordinates, $a, b \in \mathbb{R}^+$ represents the major and minor axis and $d \in \mathbb{R}^+$ stands for the curvature degree. A superellipse is parameter-compact with as few as 3 parameters, naturally preserves self-intersection, and offers a flexible representation from rectangular to star shapes. It offer a straightforward way to model essential shapes such as squares and circles, striking a balance between parameter simplicity and representation versatility

For the 3D sweeping axis, our methodology employs a third-order B-spline characterized by $n$ control points $\{c_1, \cdots, c_n\}\in \mathbb{R}^3$, ensuring a flexible yet precise control over the shape's curvature. The spline is parametrized over a clamped knot vector, guaranteeing that the B-spline's trajectory starts at $c_1$ and concludes at $c_n$. The spline curvature is modulated by the intermediate control points positions. In scenarios where these control points align collinearly, the sweep axis simplifies to a straight line, accommodating the fundamental sketch-and-extrude cases.

\begin{figure}
    \centering
    \includegraphics[width=\linewidth]{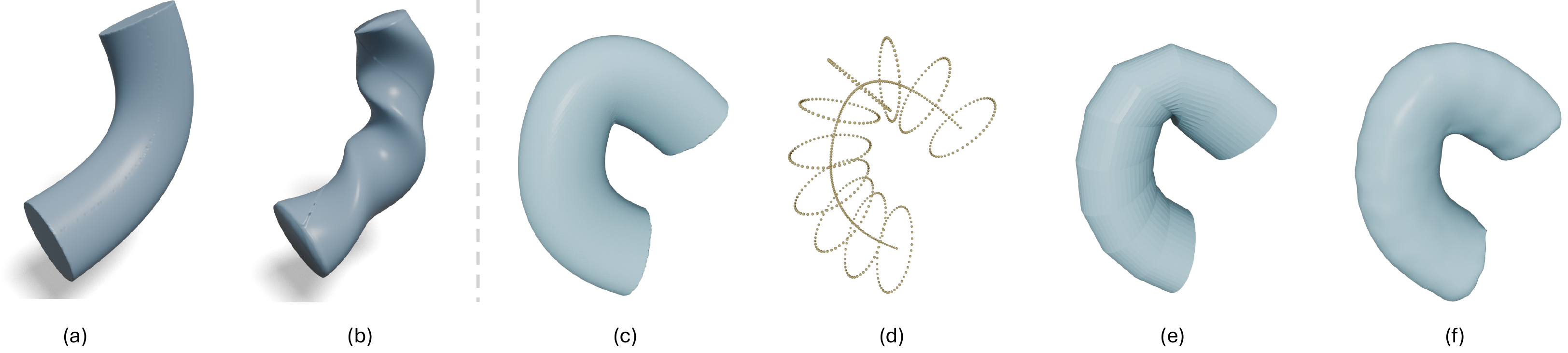}
    \caption{Sweep surfaces visualised under various conditions. (a): Sweep surface produced with parallel-transportation frame regularization. (b): Sweep surface produced with free profile rotations. (c): Reference sweep surface produced by off-the-shelf sweeper. (d) Point cloud sampled from sweep surfaces for neural sweeper input. (e) Sweep surface produced by naively interpolating profile stamps, notice the sharp crease and aliasing effect at high curvature regions. (f) Sweep surface produced by neural sweeper with the input point cloud in (d).}
    \label{fig:sv-vis}
\end{figure}

To address the inherent rotation ambiguity of the profile during sweeping, we adopt the parallel transport frames to regularise the sweeping motion. \cref{fig:sv-vis} (a), (b) show two sweep surfaces produced with the same profile and axis but with different sweeping motions. The convention of the sweep motion is aligning the profile normal (z-axis) with the sweeping trajectory tangent, which still leaves the $x-y$ axis of the profile indeterminate. The parallel transport coordinate frame establishes consistent local coordinate frames along the B-spline curve, eliminating the ambiguity in profile orientations. This method is particularly effective in maintaining consistent and deterministic sweeping motion, while being robust against extreme curvature scenarios, such as inflection points. As a result, the translation and rotation of the profile are uniquely derived along the sweeping motion, enhancing the model's precision and reliability.

The last component, scaling function $f$, dynamically adjusts the profile scale along the sweep. Defined as  $f(t):[0,1] \rightarrow \mathbb{R^+}$ for a sweeping axis $s(t):[0,1]\rightarrow \mathbb{R}^3$, it ensures the profile is appropriately scaled at each point $s(t)$. To achieve a smooth and continuous sweep, $f$ needs to be strictly $C^0$ continuous. We choose degree $k$ polynomials with a fixed constant term to formulate a scaling function, offering both effective scaling behaviour and a compact parametrization. Collectively, a sweep surface primitive is uniquely defined by: 
\begin{equation}
    S = [c_1, \cdots, c_n, a, b, d, f_1, \cdots, f_{k}] \in \mathbb{R}^{3n+k+3}
\end{equation}
\Cref{fig: superellipse} showcases sweep surfaces with various profiles, sweeping axes, and in different combinations with the scaling function. 

\subsection{Neural Sweeper}
The integration of sweep surface primitives into a learning framework is complicated by the lack of a differentiable method to generate these primitives from their defining parameters. Traditional approaches for creating sweep surfaces involve densely sampling profile frames along the sweeping axis. This sampling is followed by one of two methods: connecting adjacent profile points to construct an explicit swept volume mesh \cite{trimesh} or employing numerical continuation to compute an implicit swept volume field as the profile traverses the sweeping axis \cite{sellan2021swept}. Unfortunately, both methods present integration challenges within a learning context due to their non-differentiable nature. Moreover, a direct analytical approach, involving dense sampling and occupancy interpolation between profile frames, incurs significant computational costs and is susceptible to aliasing effects, particularly with sharply curved profiles. An example is provided in \cref{fig:sv-vis} (e), where the sweep surface produced from interpolation shows visible coarse granularity and sharp creases at high curvature regions.

In response to these challenges, we introduce the concept of a neural sweeper, a differentiable surrogate for sweep surface generation. This approach begins with the use of a differentiable sampler to collect sweep surface key points. The key points are collected by sampling points along the sweeping axis, and 3D profile slices by sparsely transforming the 2D profile into 3D space based on curve coordinate frames (\cref{fig:sv-vis} (d)). These sampled points are then processed by the neural sweeper to compute the corresponding implicit field. For model training, we generated a dataset comprising sweep surface samples with varied parameters, and applied the existing point-cloud-to-implicit-field model, POCO~\cite{boulch2022poco}, as the backbone model. Once the model is trained, we freeze the neural sweeper and plug it into \papername to facilitate subsequent outer scope training sessions. The selected architecture of the neural sweeper must support smooth gradient flow across interfaces with other components. We observed that the use of a discretized feature grid and stochastic sampling can impede gradient flow, potentially obstructing the training process of the larger framework. However, the POCO model is particularly advantageous in this context, capable of capturing detailed implicit shape features while maintaining compatibility with back-propagation techniques, thereby serving as an effective submodule within our proposed methodology.

\subsection{Training and Inference}
\papername implements a two-phase training strategy: initially, the neural sweeper is trained, followed by the comprehensive training of \papername itself. The training of the neural sweeper utilizes Binary Cross-Entropy loss to evaluate the accuracy of the predicted occupancy fields for sweep surfaces. Training \papername model involves a blend of reconstruction loss, axis loss, overlap loss, and parsimony loss to optimize shape parsing.

With $K$ sweep surfaces predicted, the neural sweeper generates occupancy fields ${O_1, \cdots, O_K}$. From these, \papername selects a subset of primitives $p_1, \cdots, p_q$ (where $q \leq K$) to construct the final shape. Testing points $T$ spread throughout the 3D space are used to calculate the reconstruction loss, which is the mean squared error between the ground truth occupancy field and the assembled occupancy field. The latter is derived using the Boltzmann operator with a sharpness parameter $\alpha 
\in \mathbb{R}$:
\begin{equation}
\mathcal{L}_{recon} = \mathbb{E}_{t\sim T}\left [ \left \lVert  O_{GT}(t) - \frac{\sum_{i=1}^{q}O_i(t)e^{\alpha O_i(t)}}{\sum_{i=1}^{q}e^{\alpha O_i(t)}} \right \rVert^2_2 \right ].
\end{equation}
Furthermore, to ensure a parsimonious representation, \papername minimizes the use of primitives through both overlap loss and parsimony loss. The overlap loss penalizes excessive overlapping among sweep surface primitives beyond a threshold $\beta$:
\begin{equation}
\mathcal{L}_{ol} = \mathbb{E}_{t\sim T}\left [\min(\sum_{i=1}^{q}O_i(t) - \beta, 0) \right].
\end{equation}
The parsimony loss, encouraging minimal primitive usage, is represented as a sublinear function of the count of selected primitives:
\begin{equation}
\mathcal{L}_{pars}=\sqrt{q}.
\end{equation}
Axis loss is introduced to guide the prediction of sweeping axes, aligning them closely with the object's medial axis. This is crucial as unsupervised learning of sweep axes is inherently ambiguous due to the multitude of possible profile-axis combinations that can generate the same object. The learning is regularized by ensuring the predicted sweeping axes encompass the object's medial axis, quantified by the chamfer distance between the medial axis points $M={m_i}$ and the points sampled from selected sweeping axis $S = \cup_{i=1}^{q}\{s_{ij}\}$:
\begin{equation}
\mathcal{L}_{axis} = \mathbb{E}_{m\sim M}\left[ \min_{s \in S} dist(m, s)\right].
\end{equation}
The overall \papername loss function is defined as:
\begin{equation}
\mathcal{L} = \lambda_1 \mathcal{L}_{recon} + \lambda_2\mathcal{L}_{ol} + \lambda_3\mathcal{L}_{pars} + \lambda_4 \mathcal{L}_{axis}.
\end{equation}
Before the training starts, the sweep surface primitives are initialized regarding the medial axis for a warm start. At inference time, we use off-the-shelf sweepers \cite{sellan2021swept} to create explicit sweep surface primitives from the predicted parameters, bypassing the neural sweeper to improve speed and accuracy. Empirically, we set $\lambda_1=12$, $\alpha=40$, $\lambda_2=6$, $\beta=0.8 \cdot K$, $\lambda_3= 0.3 K / 8$ and $\lambda_4 = 5$. More details about the hyper-parameter setting and training practice can be found in the supplemental material.

\section{Results}

In this section, we begin by offering comprehensive insights into our datasets and implementation methodologies. Subsequently, we present both quantitative metrics and qualitative observations of our method compared to alternative approaches. We also provide ablation studies to justify our design choices. More results and ablations will be provided in the supplementary material. 
\subsection{Dataset and Implementation Details}

We conduct experiments over two datasets, a custom \textit{GC-Object} dataset containing 50 models sourced from prior works \cite{zhou2015generalized, sawdayee2023orex} and internet; and quadrupeds dataset~\cite{tulsiani2017learning,xu2023animal3d} with 124 animal shapes. The data are preprocessed following the scheme of CAPRI-NET~\cite{yu2022capri}.

We showcase the parametric lightness and sweep-versatility properties of sweep surfaces. To this end, we compare SweepNet with several baseline models: two primitive-fitting shape abstraction methods using superquadrics (SQ) \cite{paschalidou2019superquadrics} and cuboids (Cuboid) \cite{yang2021unsupervised}, one network using sketch-and-extrude primitives with neural profiles (SECAD-Net) \cite{li2023secad}, one network using sketch-and-extrude primitives with CSG operations (ExtrudeNet) \cite{ren2022extrudenet} and one network using geons with CSG operations (UCSG) \cite{kania2020ucsg}. In addition, we provide insight on SweepNet with point cloud input modality by switching the encoder to DGCNN~\cite{phan2018dgcnn} module, denoted as $\text{Ours}_\text{pcd}$, showcasing the flexibility of our pipeline.
 
Our intention with this comparison is to demonstrate that, within a comparable training period, SweepNet produces superior results with minimal training iterations. Sweep surfaces exhibit greater expressiveness and versatility when dealing with curvy objects compared to conventional parametric and sketch-and-extrude primitives. This highlights the necessity of introducing sweep surfaces as a new primitive for shape abstraction

Since our model works on a per-shape basis, we empirically adapt the training scheme for the baseline models to accommodate single-shape fitting. We prioritize using the default setups for each baseline model. If they do not converge well within this setup, we increase the training iterations and/or enhance supervision signals, capped at a maximum of 10 minutes per shape training cost on an NVIDIA RTX4090 GPU.

For SQ, we train the model on each input shape for 4,000 iterations. For Cuboid, we train the model on each input shape for 10,000 iterations. For SECAD-Net, we pretrain the model on the entire dataset for 1,000 epochs, followed by fine-tuning on each model for another 2,000 iterations before inference. ExtrudeNet and UCSG require longer training epochs to learn CSG operations. For ExtrudeNet, we replace the voxel input with a point cloud of 32,764 points and provide 100,000 occupancy points for supervision (3.05 $\times$ our input). ExtrudeNet is trained for 60,000 iterations, and UCSG is trained for 40,000 iterations. For SweepNet, we train the model on each input shape for 2,000 iterations without any pre-training. All models are trained with a maximum of eight primitives.

\subsection{Quantitative Comparisons}
We present quantitative measurements obtained for Chamfer-Distance (CD), Volumetric Intersection over Union (IoU), and F-score with an accuracy threshold of 0.05 (F1)~\cite{tatarchenko2019single}. 

The detailed quantitative results are presented in Table~\ref{tab:GC} for the GC-Object dataset and in Table~\ref{tab:animal} for quadrupeds. Across all three metrics provided for the GC-Object dataset, our method outperforms others. In the quadrupeds dataset, our method demonstrates superior performance in all metrics compared to other methods, except for IoU against SQ \cite{paschalidou2019superquadrics}, where our method falls slightly short.

%  \begin{figure}
%     \centering
%     \includegraphics[width=\linewidth]{figures/animal3D_qualitative_placeholder.png}
%     \caption{Test result on animal3D with comparisons to SECAD, ExtrudeNet and SQ}
%     \label{fig:enter-label}
% \end{figure}

\begin{table}[ht]
\centering
\caption{Quantitative evaluations on GC-Object dataset. Our method outperforms other alternatives in all the metrics included. }
\resizebox{\linewidth}{!}{
\begin{tabularx}{\linewidth}{|>{\centering\arraybackslash}X|>{\centering\arraybackslash}X|>{\centering\arraybackslash}X|>{\centering\arraybackslash}X|>{\centering\arraybackslash}X|>{\centering\arraybackslash}X|>{\centering\arraybackslash}X|>{\centering\arraybackslash}X|}
\toprule
                          & EN & UCSG & SECAD & Cuboid & SQ & Ours & $\text{Ours}_{\text{pcd}}$ \\ \midrule
IoU $\uparrow$ & 0.420 & 0.554 & 0.584 & 0.343 & 0.597 & \cellcolor{brightyellow}0.608 & \cellcolor{brightcyan}0.620  \\ \midrule
CD $\downarrow$  & 0.0481 & 0.0170 & 0.0179 & 0.0280 & 0.0182 & \cellcolor{brightcyan}0.0168  & \cellcolor{brightcyan}0.0168  \\ \midrule
F1 $\uparrow$  & 0.652 & 0.947 & 0.967 & 0.871 & 0.977 & \cellcolor{brightyellow}0.984  & \cellcolor{brightcyan}0.985 \\ \bottomrule
\end{tabularx}
}
\label{tab:GC}
\vspace{-20pt}
\end{table}

\begin{table}[ht]
\centering
\caption{Quantitative evaluations on quadrupeds. Our method with voxel input slightly falls short of SQ~\cite{paschalidou2019superquadrics} while leading with point cloud input.}
\resizebox{\linewidth}{!}{
\begin{tabularx}{\linewidth}{|>{\centering\arraybackslash}X|>{\centering\arraybackslash}X|>{\centering\arraybackslash}X|>{\centering\arraybackslash}X|>{\centering\arraybackslash}X|>{\centering\arraybackslash}X|>{\centering\arraybackslash}X|>{\centering\arraybackslash}X|}
\toprule
                          & EN & UCSG & SECAD & Cuboid & SQ & Ours & $\text{Ours}_{\text{pcd}}$ \\ \midrule
IoU $\uparrow$ & 0.197 & 0.436 & 0.270 & 0.259 & \cellcolor{brightyellow}0.512 & 0.482 & \cellcolor{brightcyan}0.562  \\ \midrule
CD $\downarrow$  & 0.0668 & 0.0177 & 0.0224 & 0.0313 & 0.0179 & \cellcolor{brightyellow}0.0176  & \cellcolor{brightcyan}0.0168  \\ \midrule
F1 $\uparrow$  & 0.458 & 0.952 & 0.932 & 0.840 & 0.964 & \cellcolor{brightyellow}0.967  & \cellcolor{brightcyan}0.989 \\ \bottomrule
\end{tabularx}
}
\label{tab:animal}
\vspace{-20pt}
\end{table}

\subsection{Qualitative Comparisons}

\begin{figure}
    \centering
    \includegraphics[width=\linewidth]{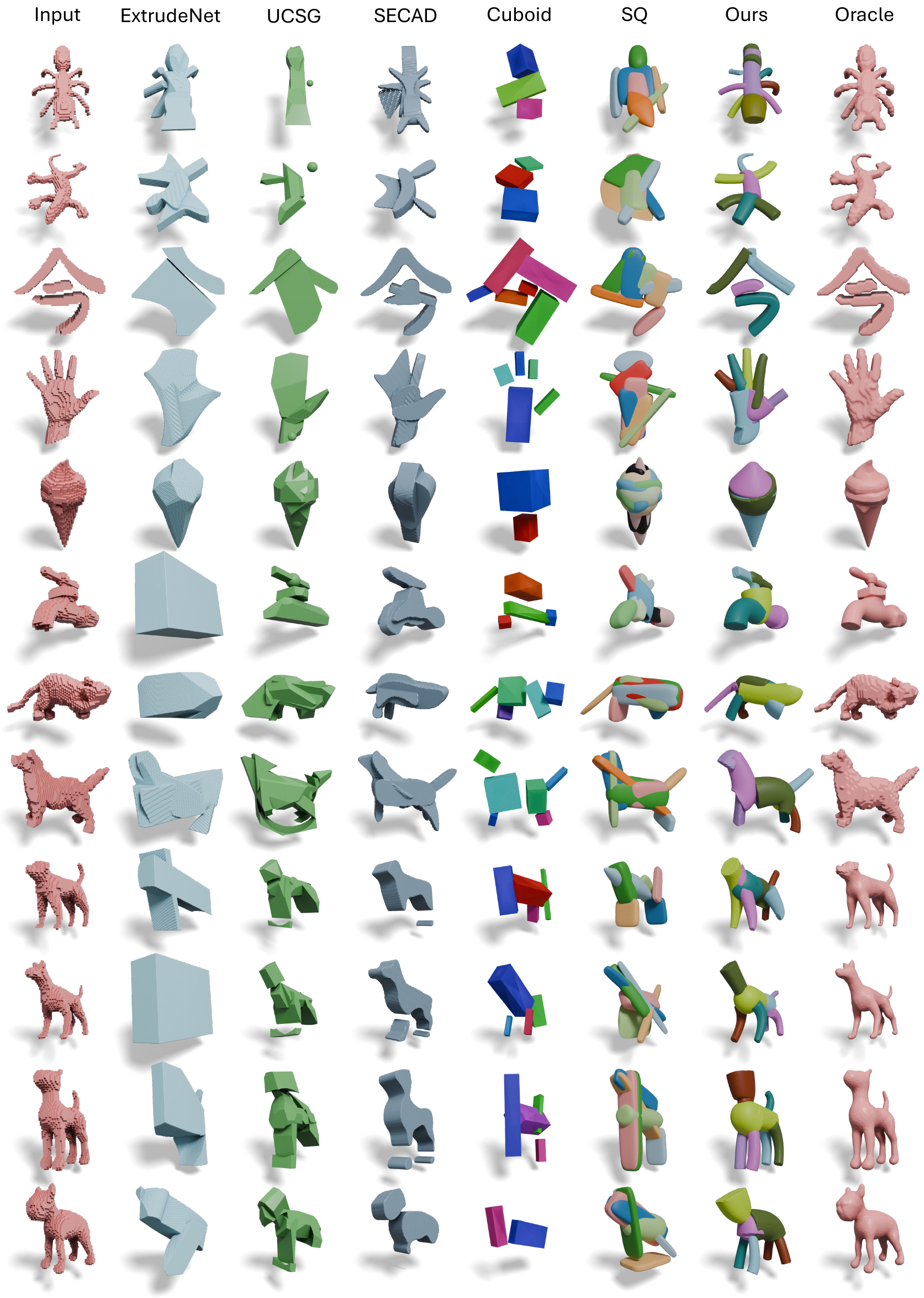}
    \caption{Qualitative comparison among ExtrudeNet~\cite{ren2022extrudenet}, UCSG~\cite{kania2020ucsg}, SECAD-Net~\cite{li2023secad}, Cuboid~\cite{yang2021unsupervised}, SQ~\cite{paschalidou2019superquadrics} and our method. Models abstracted by our method requires less primitives while better representing curvy geometric features.}
    \label{fig:qual_1}
\end{figure}

The qualitative results are demonstrated in \cref{fig:qual_1}, it can be seen that sweep surfaces, with appropriate scaling, have superior expressiveness. Sweep elements like curvy-linear limbs can be compactly represented by sweep surfaces using a single primitive, whereas other baseline primitives require multiple components for approximation. By leveraging the sweeping axis from straight lines to curves, sweep surfaces faithfully represent tubular parts such as ant legs which are challenging for sketch-and-extrude. Additionally, the scaling function enhances the versatility of sweep surfaces, effectively capturing shapes such as the gradually thinning gecko tail and the cone shape in the icecream.

\subsection{Ablation Studies}

\begin{figure}
    \centering
    \includegraphics[width=\linewidth]{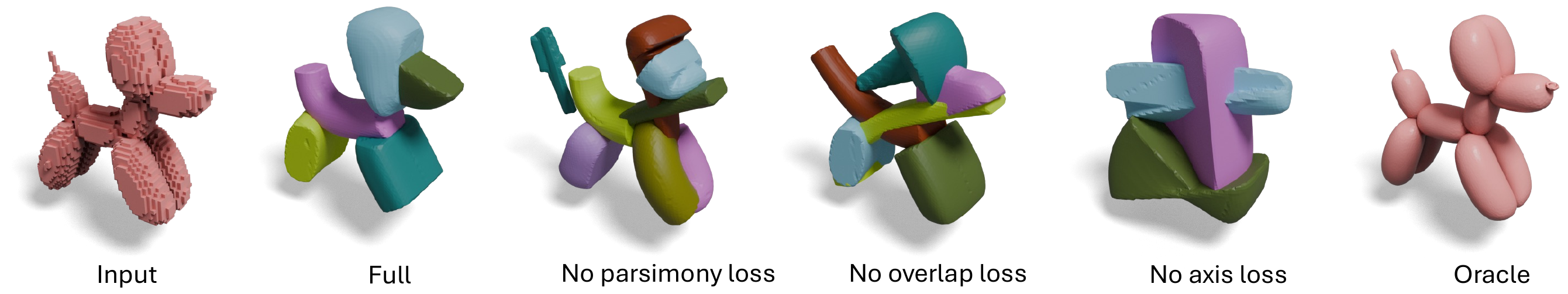}
    \caption{Shape abstraction results with various loss settings. Our model performs the best with all loss functions equipped.}
    \label{fig:loss_ablate}
\end{figure}

We conducted an extensive ablation study on our total loss designs, deactivating each loss component one at a time. In each experimental setting, we trained our model for a fixed iteration of 1,000 and simultaneously assessed the convergence speed. The qualitative outcomes are depicted in \cref{fig:loss_ablate}.

As illustrated in \cref{fig:loss_ablate}, employing the full loss incorporating all four components yields results that closely mirror the input. When $\mathcal{L}_{pars}$ is disabled, there is a noticeable increase in the utilization of primitives to compensate for fine-scale details. Similarly, omitting the overlap loss $\mathcal{L}_{ol}$ leads to the emergence of undesired overlaps, particularly evident in the body of the dog. Furthermore, excluding the axis loss $\mathcal{L}_{axis}$ results in a loss of fidelity in preserving the curvy shape of the dog, yielding a more cumbersome appearance. These observations underscore the significance of each component within our loss framework. Additional results can be found in the supplementary material.

\begin{figure}[t]
    \centering
    \includegraphics[width=0.8\linewidth]{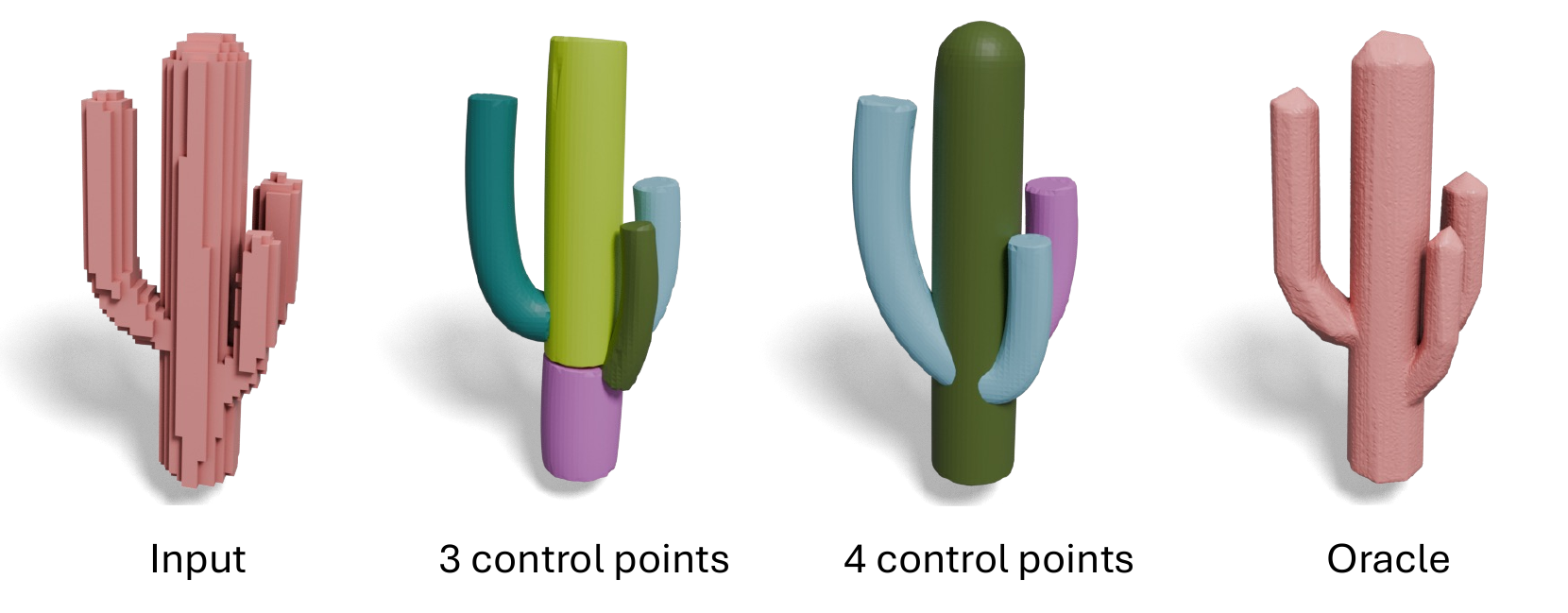}
    \caption{Ablation study showing how the number of control points affects the abstraction results. While our default is to have 3 control points for our B-spline axis, our method performs well with 4 control points too. This shows that our method is relatively robust against the number of axis control points. }
    \label{fig:cp-ablation}
    \vspace{-20pt}
\end{figure}

We additionally assess the impact of the parametric complexity of sweep surfaces by increasing the number of control points from 3 to 4. A qualitative example is demonstrated in \cref{fig:cp-ablation}. With both options, the model can make a reasonable abstraction of the input shape.
Although our standard configuration utilizes 3 control points for our B-spline axis, our method also performs effectively with 4 control points. This indicates the relative robustness of our method against variations in the number of axis control points.

\begin{figure}
	\centering
	\includegraphics[width=\linewidth]{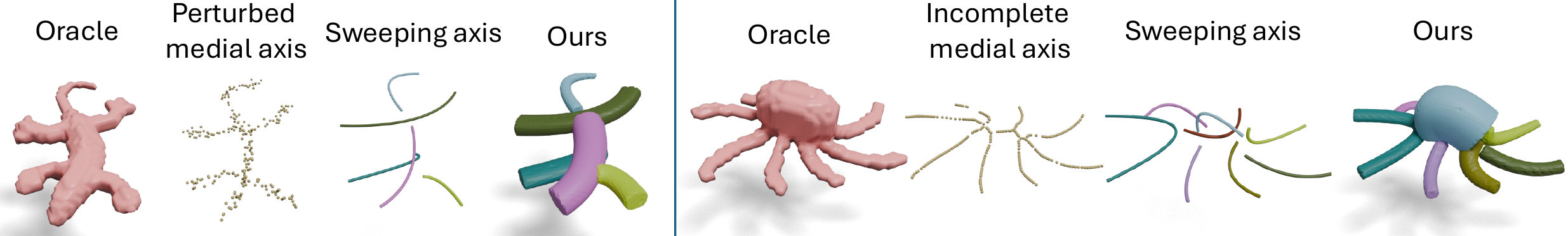}
	\caption{Shape abstraction results with noisy or incomplete medial axis guidance.}
	\label{fig:axis_ablation}	
\end{figure}

Lastly, we conduct a sensitivity test on the medial axis. The medial axis is a crucial component in SweepNet, leading to a faster and more rational fitting of sweep surfaces. Despite the reliance of SweepNet on this skeletal prior, our method exhibits a certain degree of robustness against noisy medial axes. We showcase two examples in \Cref{fig:axis_ablation}. In the first example, we inject Gaussian noise with a standard deviation of 0.01 to the extracted gecko medial axis. In the second example, the extracted medial axis of the octopus is incomplete, missing the head. In both cases, SweepNet can compensate for the faulty part and produce reasonable abstracted results.
 
\vspace{-2em}
\subsection{Editablity}
\begin{figure}
    \centering
    \includegraphics[width=\linewidth]{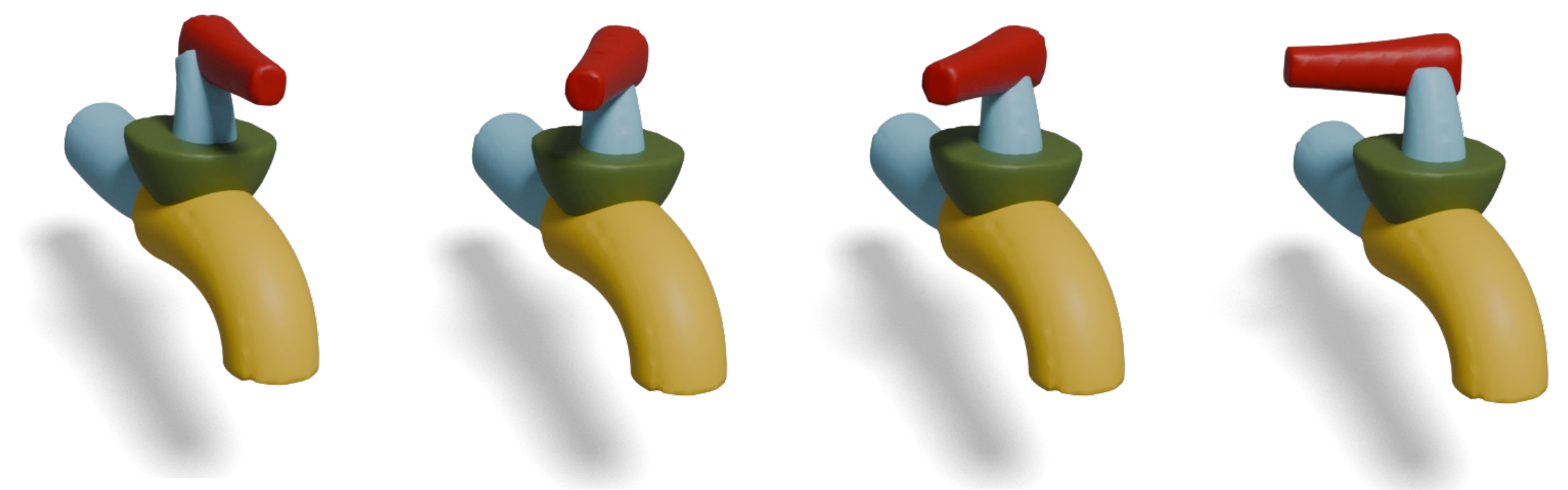}
    \caption{Primitive editing example of spinning faucet valve by altering sweep surface parameters.}
    \label{fig:edit-sample}
    \vspace{-20pt}
\end{figure}

In this section, we illustrate the flexibility of parametric sweep surfaces through post-creation editing. We present a case study in \cref{fig:edit-sample}, where the faucet valve is rotated 90 degrees clockwise by editing the associated primitive parameters. This is achieved by applying an affine transformation to the sweeping axis control point coordinates, demanding a change to only 9 float numbers. We provide additional edit examples in the supplementary material.

\section{Conclusion, Limitations, and Future Work}
\label{sec:conclusion}
\begin{figure}
     \centering
     \includegraphics[width=\linewidth]{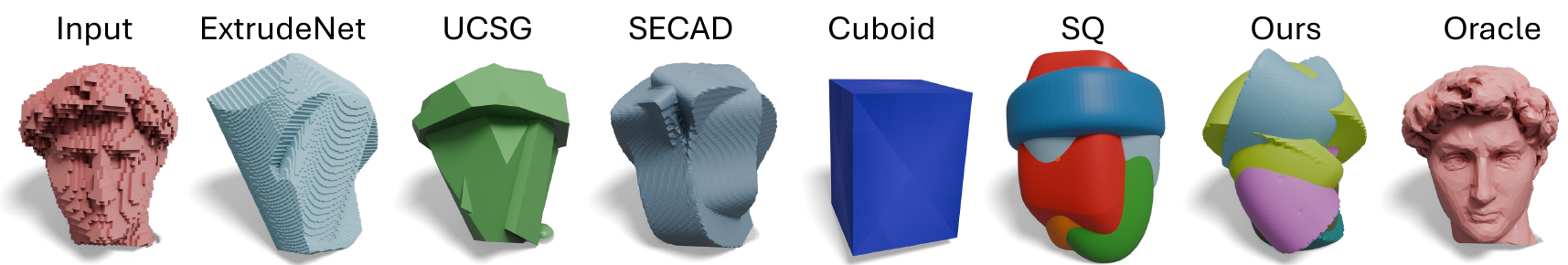}
     \caption{Failure cases of shape abstraction methods.}
     \label{fig:failurecase}
     \vspace{-10pt}
 \end{figure}

 Our method faces some limitations and future directions can be explored. The current model falls short in representing high-porosity or overly thin objects with solid sweep surfaces. Moreover, shape abstraction tends to struggle when dealing with complex models containing numerous intricate details (refer to Fig.~\ref{fig:failurecase}). Our method performs optimally when the provided model includes sweep elements. If the model lacks such elements, our method may not achieve the most favourable outcome. Hence, an intriguing area for exploration lies in integrating neural sweepers with other types of primitives within more complex systems (e.g., CSG techniques) to capture geometric intricacies while maintaining compactness. Also, currently \papername fits to each model individually, which can encounter local optimums at different initialization, future research can be done to extend this work for a generalizable shape abstraction model.
\vspace{-10pt}
     
%\section{Conclusion}
In this paper, we presented \papername, a method designed for shape abstraction through the utilization of sweep surfaces. Our approach introduces a novel and compact parameterization that facilitates intuitive editing and effectively retains shape details. The integration of neural sweepers introduces a new way to incorporate challenging primitives into shape abstraction tasks. Neural sweepers can be seamlessly plugged and played in other deep learning networks for sweep surface production or tailored to tackle other complex geometric primitives, providing a versatile tool for advancing shape abstraction techniques. Collectively, our model showcases its ability to accurately predict shape abstractions via sweep surfaces without the need for supervision. We have also demonstrated the superiority of our approach over conventional methods in shape abstraction targeting on curvy-feature objects. In conclusion, SweepNet is a step further in 3D shape abstraction, combining the strengths of sweep surfaces with efficient parameterization and dynamic scaling. While there are limitations to address, the potential for future enhancements is vast.

\section*{Acknowledgement}
We thank Cody Reading for the insightful discussions and Ruiqi Wang for the help with figure preparations.
% \clearpage
% \bibliography{egbib}

\appendix

\clearpage
\setcounter{page}{1}
\title{SweepNet Supplementary Material} 

\titlerunning{SweepNet}

\author{Mingrui Zhao\inst{1} \and
Yizhi Wang\inst{1} \and
Fenggen Yu\inst{1} \and
Changqing Zou \inst{2} \and \\
Ali Mahdavi-Amiri \inst{1}}

\authorrunning{M.~Zhao et al.}

\institute{Simon Fraser University \and
Zhejiang University\\}
\maketitle
%%%%%%%%%% Merge with supplemental materials %%%%%%%%%%
%%%%%%%%%% Prefix a "S" to all equations, figures, tables and reset the counter %%%%%%%%%%
\setcounter{equation}{0}
\setcounter{figure}{0}
\setcounter{table}{0}
\setcounter{page}{1}
% \makeatletter

\section{Neural Sweeper Training}
\begin{figure}
    \centering
    \includegraphics[width=\linewidth]{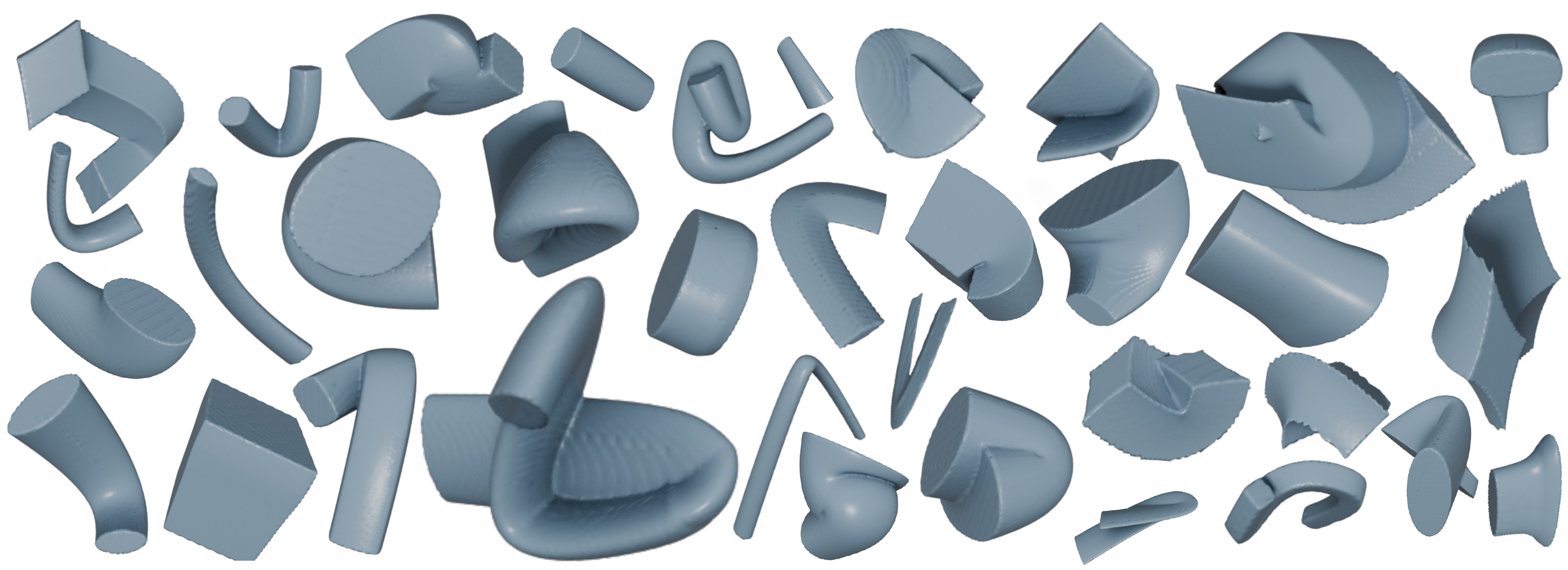}
    \caption{Sweep surfaces data samples for neural sweeper training.}
    \label{fig:sv-dataset}
\end{figure}
\vspace{-5pt}
\subsection{Data Preparation}
To fully train our neural sweeper, we create a dataset of sweep surfaces with a variety of parameters using the technique introduced by S\`ellan et al.~\cite{sellan2021swept}. \cref{fig:sv-dataset} showcases some randomly selected samples in the dataset. 
%SweepNet takes input as a point cloud of a sweep surface and predicts its SDF.
Specifically, the parameters of a sweep surface data sample are set as follows:

\paragraph{Sweeping Axis.} As mentioned in our main paper, the sweeping axis is a B-spline curve. Control points for the sweeping axis are randomly generated within the range $[-0.5, 0.5]^n$, where $n$ is the number of control points. Since the input shapes to SweepNet (our main network) are normalized within the unit cube, this range selection ensures that the sweeping axis remains within the confine, preserving the integrity of the sweep surfaces.

\paragraph{Superellipse Profile.} Parameters for the superellipse profile, including the major-minor axis and degree, are selected randomly within the ranges $[0.01, 0.5]^2$ and $[0.3, 5]$, respectively. This approach prevents the generation of overly small profiles which could potentially hinder the learning process. The degree lower bound is set to be $0.3$ to avoid extreme star-shape profile with diminishing corners.

\paragraph{Scaling Function.} We adopted quadratic function $ax^2+bx+c$ with $c=1$ for the profile scaling. This setup allows for constant scaling when $a$ and $b$ are set to zero. The parameters $a$ and $b$ are confined within the range $[-0.5,0.5]$, ensuring the scaling velocity remains within a reasonable limit.

\subsection{Point Cloud Sampling}
After obtaining the mesh of a sweep surface, we sample a point cloud from it as a training input to our neural sweeper.
For each sweep surface, we sample (a) 15 profile frames, with each consisting of 50 contour points and (b) 124 points from the sweeping axis. There are altogether $15 \times 50 + 124$ points as a point cloud fed to the neural sweeper. Traditionally, a point cloud contains only the points from the object surface. Here we include the points from the sweeping axis as auxiliary information in the network input. The sampling details are as follows:

\begin{itemize}
    \item The axis points are uniformly sampled from the B-spline curve's parameter space.
    \item The contour points are uniformly sampled using the superellipse formulation with $\theta$ ranging from $0$ to $2\pi$.
    \item The profile frames position are uniformly sampled from the spline curve in parameter space. For each profile frame position, a 2D profile is scaled with the scaling function then transformed to the corresponding position with transformation matrix calculated by the parallel-transportation frame.
\end{itemize}

\subsection{Training Strategy}
The training protocol follows the implementation of POCO~\cite{boulch2022poco}, using an Adam optimizer with a learning rate of $1\mathrm{e}-3$. After completing training, the neural sweeper is frozen and cascaded to the swept volume head in SweepNet. 
%The swept volume head output is aligned with the neural sweeper receptive field.

\section{Additional Results}
\begin{figure}
    \centering
    \includegraphics[width=\linewidth]{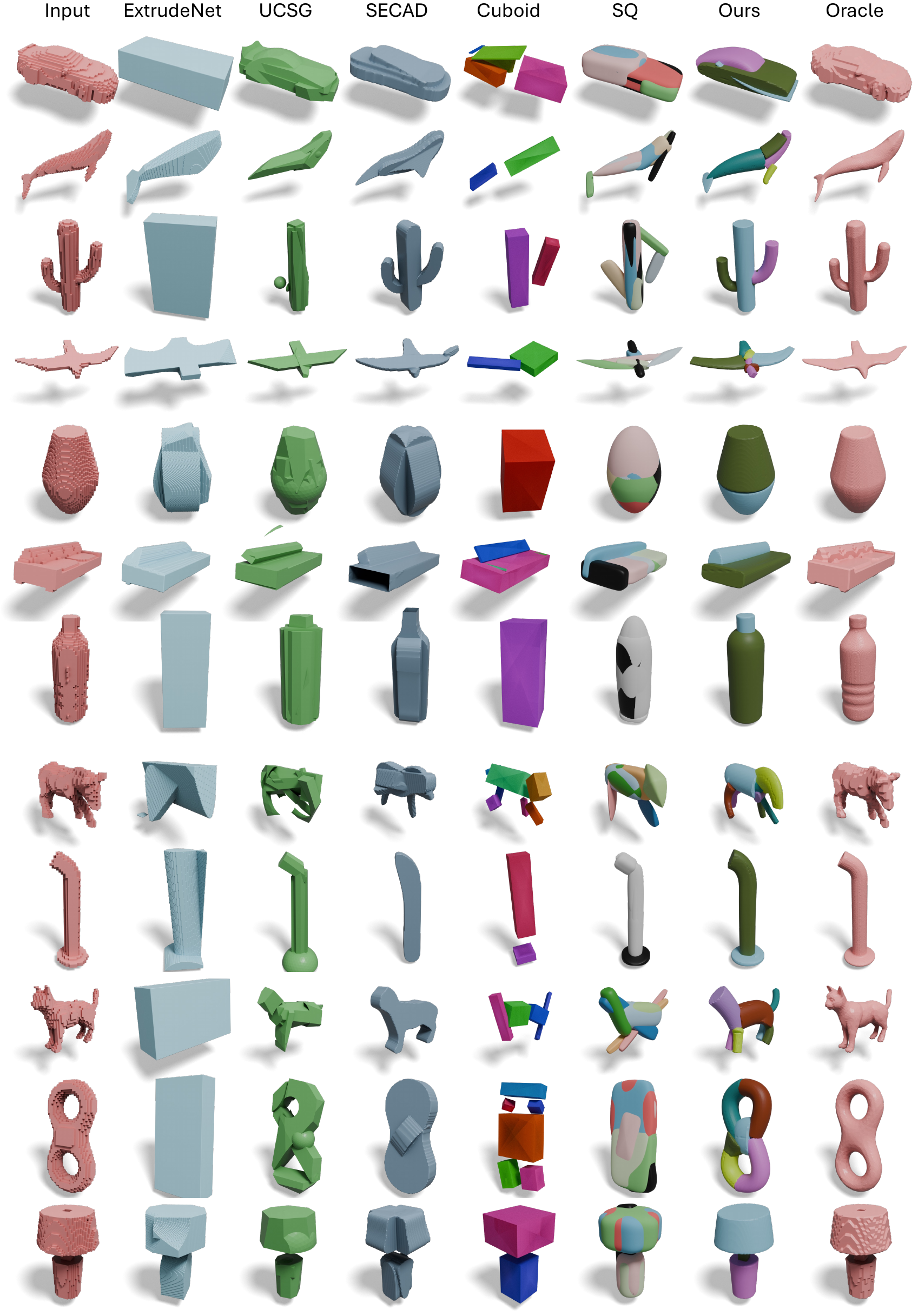}
    \caption{Additional qualitative results on GC-objects and quadrupeds datasets. Our method better captures curves.}
    \label{fig:add-qual}
\end{figure}

More visual results and comparisons are provided in Fig.~\ref{fig:add-qual}.
The results demonstrate the power of sweep surfaces in representing curved surfaces, and our proposed scaling function further enhances their expressive capabilities.

\begin{figure}
    \centering
    \includegraphics[width=0.65\linewidth]{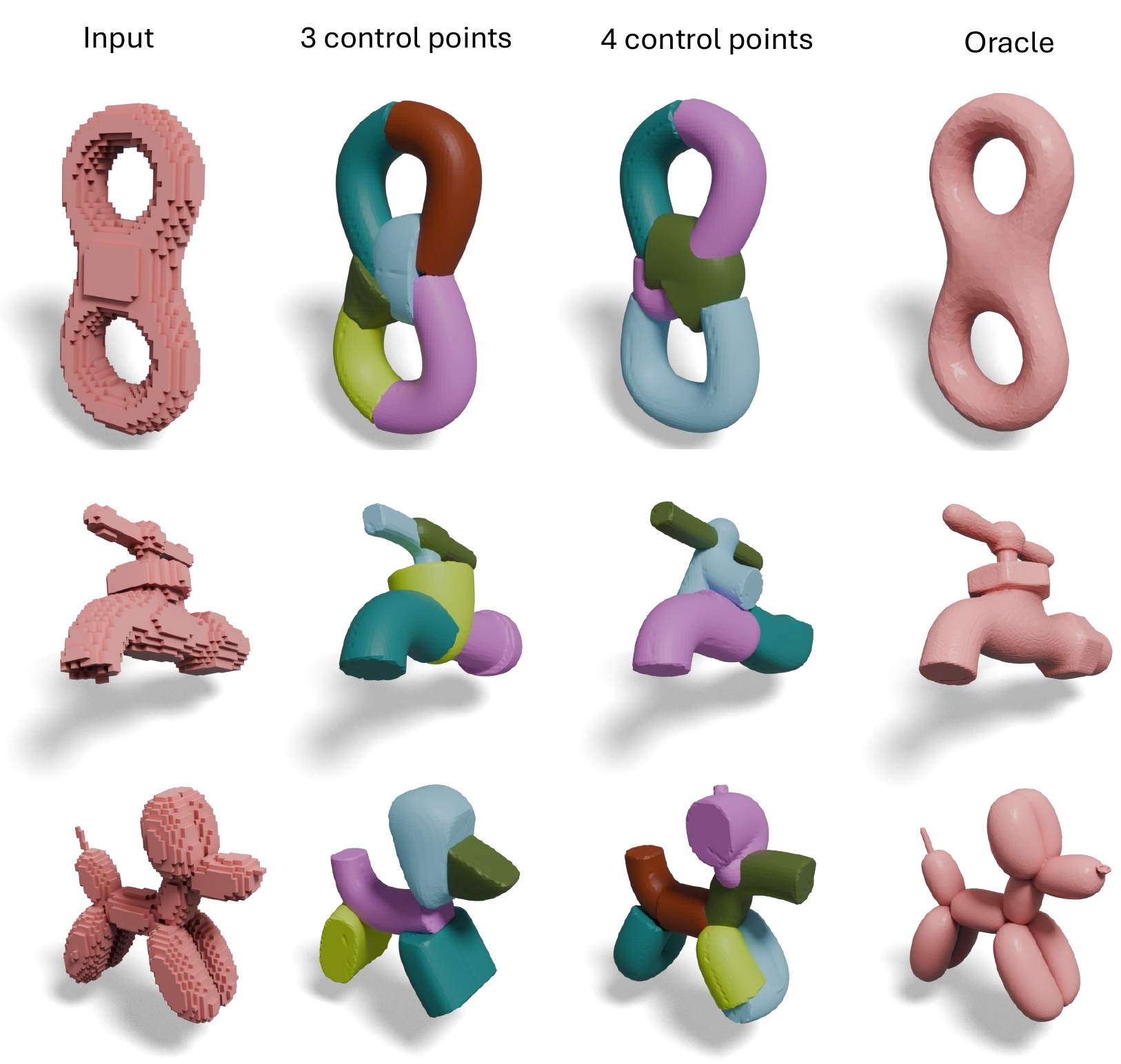}
    \caption{Shape abstraction outcomes from SweepNet utilizing sweeping axes defined by 3 and 4 control points, respectively. Primitives characterized by 3 control points are more concise, while those with 4 control points showcase more turning curves and employ fewer primitives in abstraction.}
    \label{fig:sup-cpablation}
\end{figure}

\cref{fig:sup-cpablation} provides additional examples to show how the number of control points affects the abstraction results.
A sweeping axis with more control points exhibits more curvy features, while our method remains stable in both settings.

Despite curvy-feature objects, we provide additional qualitative results of SweepNet on Thingi10K~\cite{zhou2016thingi10k} and ShapeNet~\cite{chang2015shapenet} datasets. These datasets contain many CAD-like shapes where the sweep elements are not commonly observed. Our method can provide comparable results. We would like to emphasize that no single primitive is perfect for all 3D shapes. Each primitive has its strengths and weaknesses, and the combination of various methods can often provide a more comprehensive solution. SweepNet excels in handling objects with curvy and tubular features but may not be the best fit for CAD-like shapes. Combining SweepNet with other parametric primitives can harness the strengths of each method to achieve better overall performance.

\begin{figure}
    \centering
    \includegraphics[width=\linewidth]{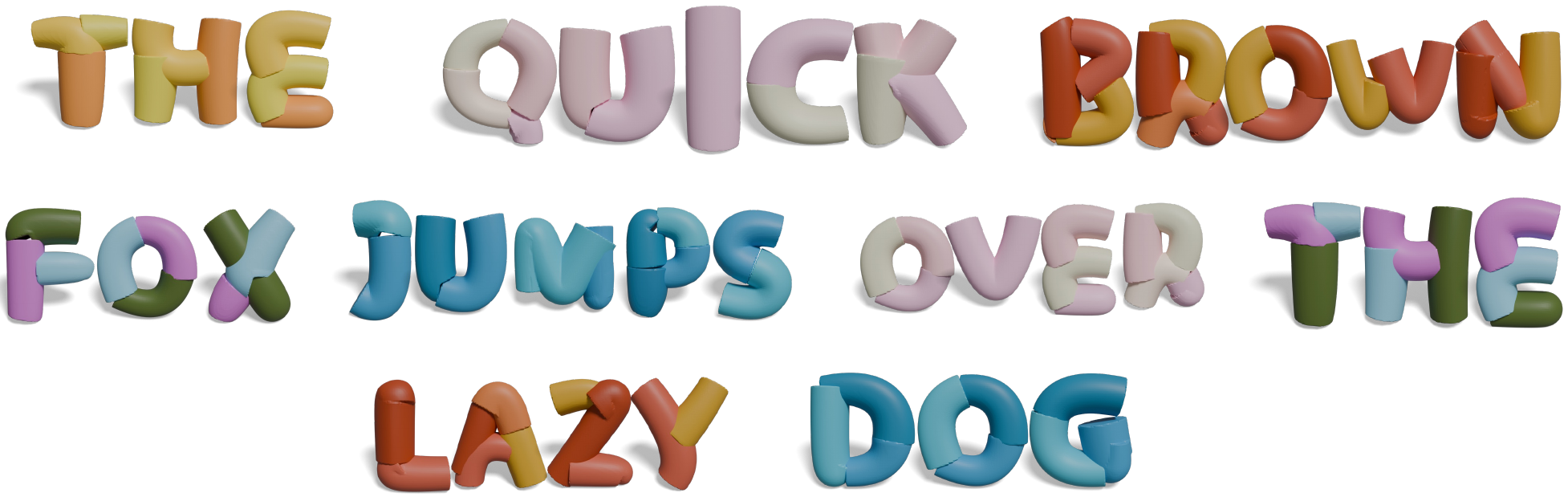}
    \caption{Representing letters with sweep surfaces produced by SweepNet.}
    \label{fig:letter}
\end{figure}

\begin{figure}
    \centering
    \includegraphics[width=\linewidth]{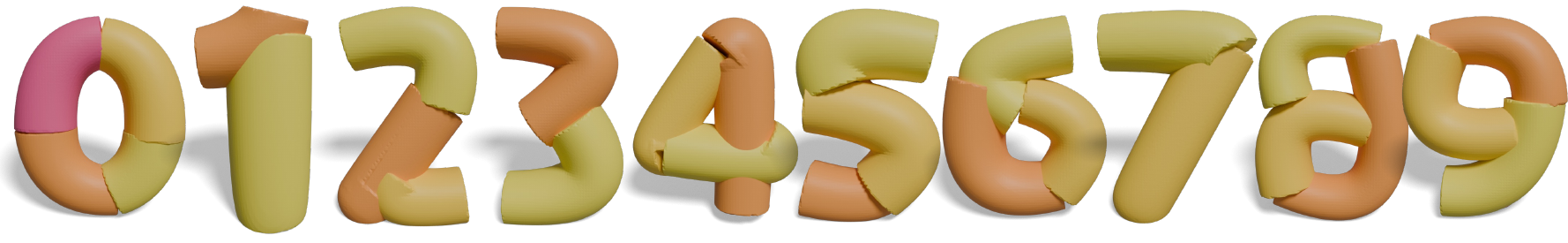}
    \caption{Representing numbers with sweep surfaces produced by SweepNet.}
    \label{fig:number}
\end{figure}

Lastly, we provide qualitative results from SweepNet in representing alphabetical letters and numbers in \cref{fig:letter,fig:number}.

\section{Loss function}
As detailed in the main paper, we propose four distinct loss functions: $\mathcal{L}_{recon}$, $\mathcal{L}_{ol}$, $\mathcal{L}_{pars}$, and $\mathcal{L}_{axis}$. This section elucidates the rationale behind the design of each loss function and delineates the methodology for tuning their respective weights.

\subsection{Reconstruction Loss Formulation}
We used the Boltzmann operator to formulate the reconstruction loss to enable a smoother gradient flow. This operator calculates the occupancy value at a test point by taking the weighted sum of the occupancy values from all present primitives, ensuring that all primitive parameters can be updated during backpropagation. The weight is determined by a biased softmax function controlled by the parameter $\alpha$. Alternatively, the argmax operator could be used to update only the primitive contributing the highest occupancy, but this method empirically slows convergence. The Boltzmann operator provides a smooth approximation of the maximum function, and its sharpness toward the true maximum can be adjusted by tuning the parameter $\alpha$.

\begin{equation}
\mathcal{L}_{recon} = \mathbb{E}_{t\sim T}\left [ \left \lVert  O_{GT}(t) - \frac{\sum_{i=1}^{q}O_i(t)e^{\alpha O_i(t)}}{\sum_{i=1}^{q}e^{\alpha O_i(t)}} \right \rVert^2_2 \right ].
\end{equation}

\subsection{Loss Tunning Strategy}
The primary loss function, $\mathcal{L}_{recon}$, measures the fidelity between the abstracted shape representations and their corresponding GT counterparts. Nevertheless, relying solely on $\mathcal{L}_{recon}$ may result in suboptimal abstraction. This is characterized by a tendency of the model to produce aggregated and cumbersome representations that merely approximate the target shape, without achieving meaningful abstraction. To address this, we introduced the overlap loss, $\mathcal{L}_{ol}$, and the parsimony loss, $\mathcal{L}_{pars}$, to encourage more parsimonious reconstructions by penalizing overlapping primitives and the excessive use of primitives. Empirical observations suggest that the overlap loss also facilitates faster convergence by introducing a repulsive force among primitives.

Furthermore, the axis loss, $\mathcal{L}_{axis}$, guides the selection of sweeping axes towards the medial axis of the target object, which can be directly extracted from the input voxel data.
%, thus eliminating the need for external information. 
Incorporating the medial axis as additional supervision significantly enhances our model's performance, rapidly narrowing the solution space and establishing a robust prior for the sweeping axis.

\begin{figure}
    \centering
    \includegraphics[width=0.95\linewidth]{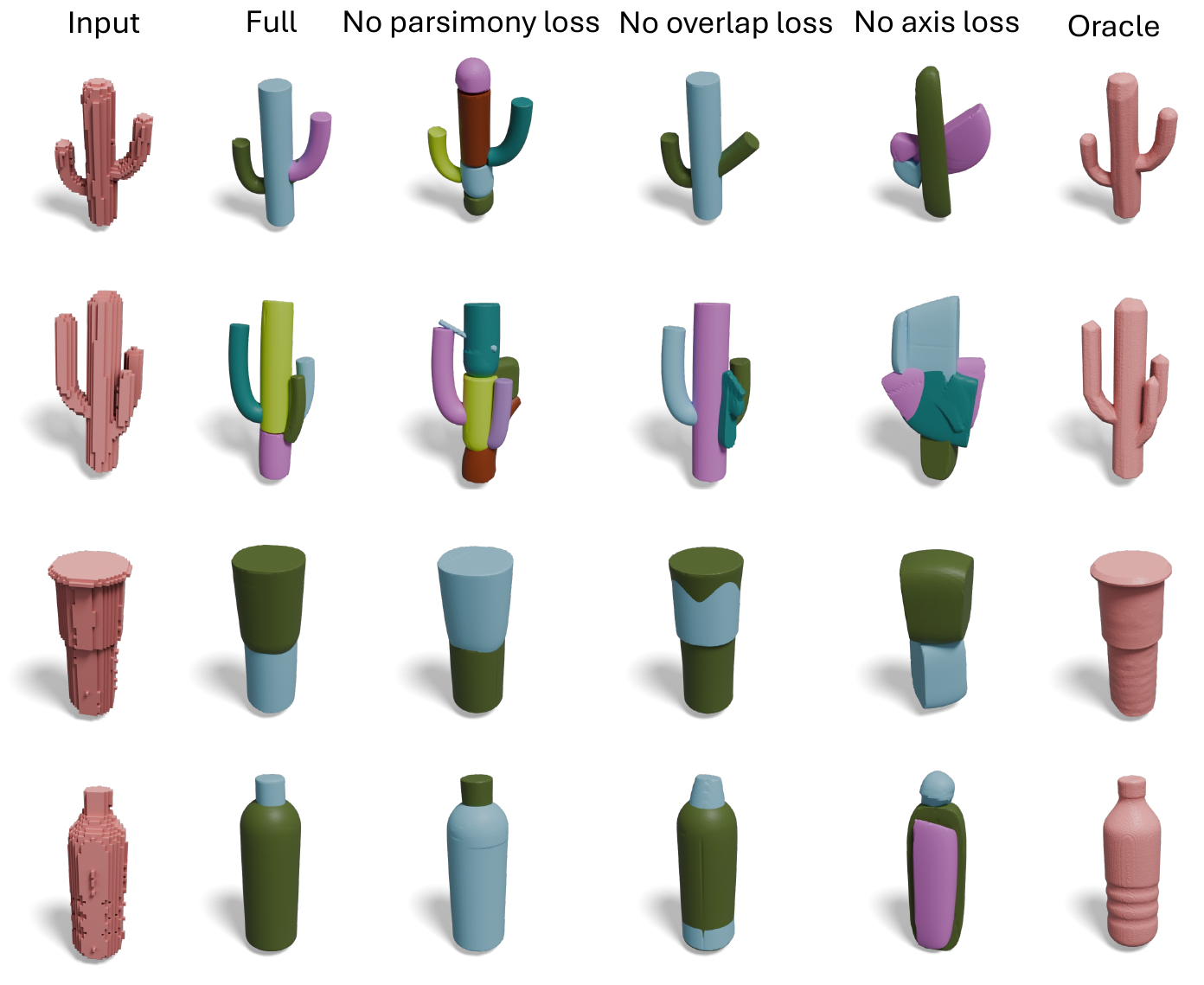}
    \caption{Ablation study on loss functions. Parsimony loss encourages minimal use of primitives, overlap loss reduces overlaps, and axis loss provides superior guidance on sweeping axes}
    \label{fig:ablation-loss}
\end{figure}

When determining the weights assigned to each loss function, we consider the following principles: The reconstruction loss has the dominant weight to ensure a faithful shape representation, followed by the overlap loss to regularize cleanliness. The parsimony loss introduces a trade-off between parsimony and fidelity, so it is set to a comparatively small value to preserve fidelity. The axis loss is mandatory but only serves as a prior. It is set with a decaying weight, with higher importance at the beginning of the training process. The impact of each loss function is shown in \cref{fig:ablation-loss}.

\section{Primitive Edits}

\begin{figure}
    \centering
    \includegraphics[width=0.8\linewidth]{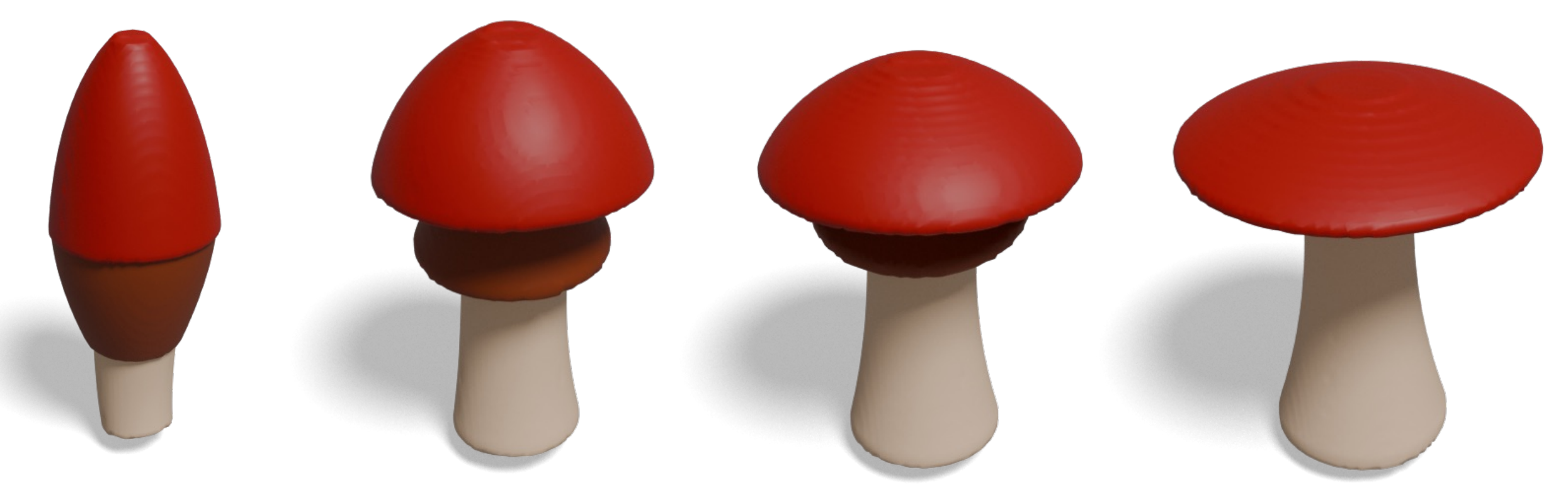}
    \caption{A light bulb model is transformed into a mushroom by manipulating sweeping axis length, adjusting the scaling function, and enlarging the profile size. Each shape is represented by three primitives, with three control points each, employing a quadratic scaling function. In total, this representation requires 42 floating-point numbers.}
    \label{fig:bulb-mushroom}
\end{figure}

\begin{figure}
    \centering
    \includegraphics[width=0.8\linewidth]{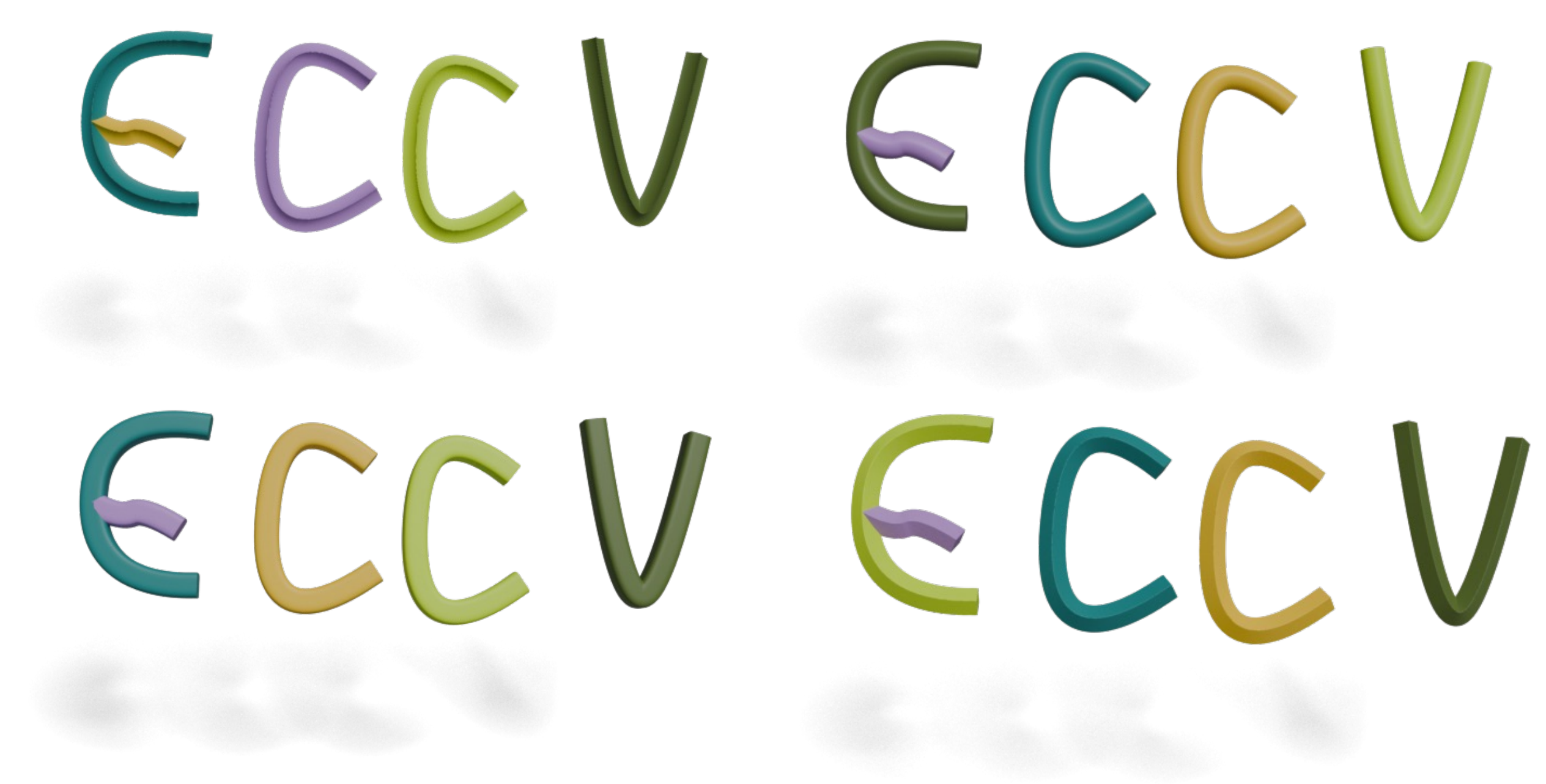}
    \caption{An ECCV 3D calligraphy undergoes variations in its brush style by adjusting the degree of the superellipse profile. This adjustment creates different profiles, including a star shape profile (top left), circular profile (top right), square profile (bottom left), and rhombus profile (bottom right), showcasing various font effects. Each shape is represented by five primitives, each with four control points and a constant scaling function, totaling 85 floating-point numbers. Only \textbf{one} float from each primitive needs to be updated to achieve this edit.}
    \label{fig:eccv-edit}
\end{figure}

\begin{figure}
    \centering
    \includegraphics[width=0.8\linewidth]{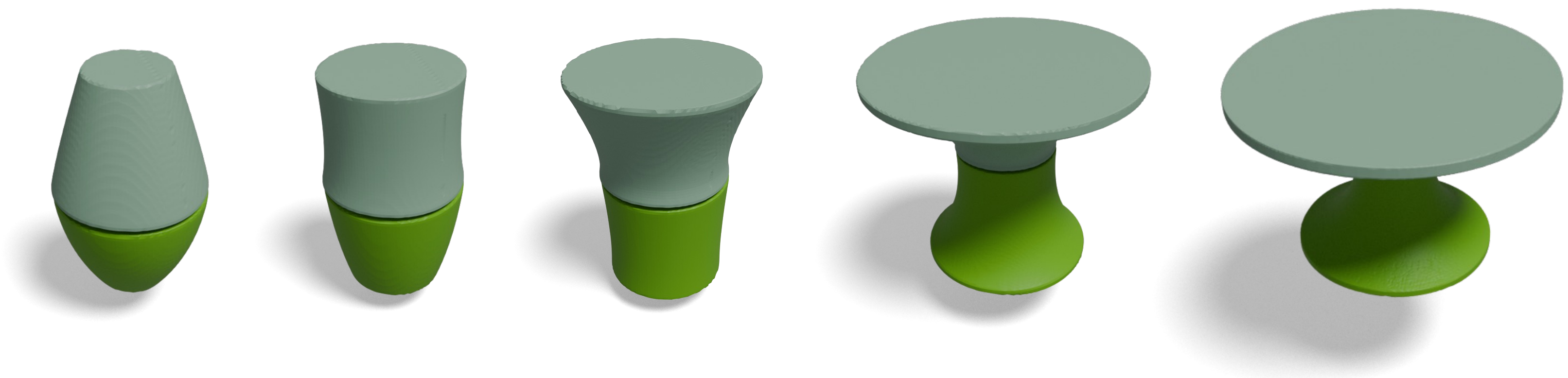}
    \caption{A vase model transforms to a table by adjusting the sweeping axis length, and scaling function coefficients. Each shape is represented by two primitives, each with three control points and a quadratic scaling function, requiring a total of 28 floating-point numbers.}
    \label{fig:faucet-edit}
\end{figure}

We showcase additional examples of primitive editing from \cref{fig:bulb-mushroom,fig:eccv-edit,fig:faucet-edit}. The abstracted shapes demonstrate versatile editing capabilities of our method by altering the profile, axis, and scaling functions of the primitives. These examples highlight the advantages of the proposed sweep surface parametrization and its flexibility.


\begin{figure}
    \centering
    \includegraphics[width=0.9\linewidth]{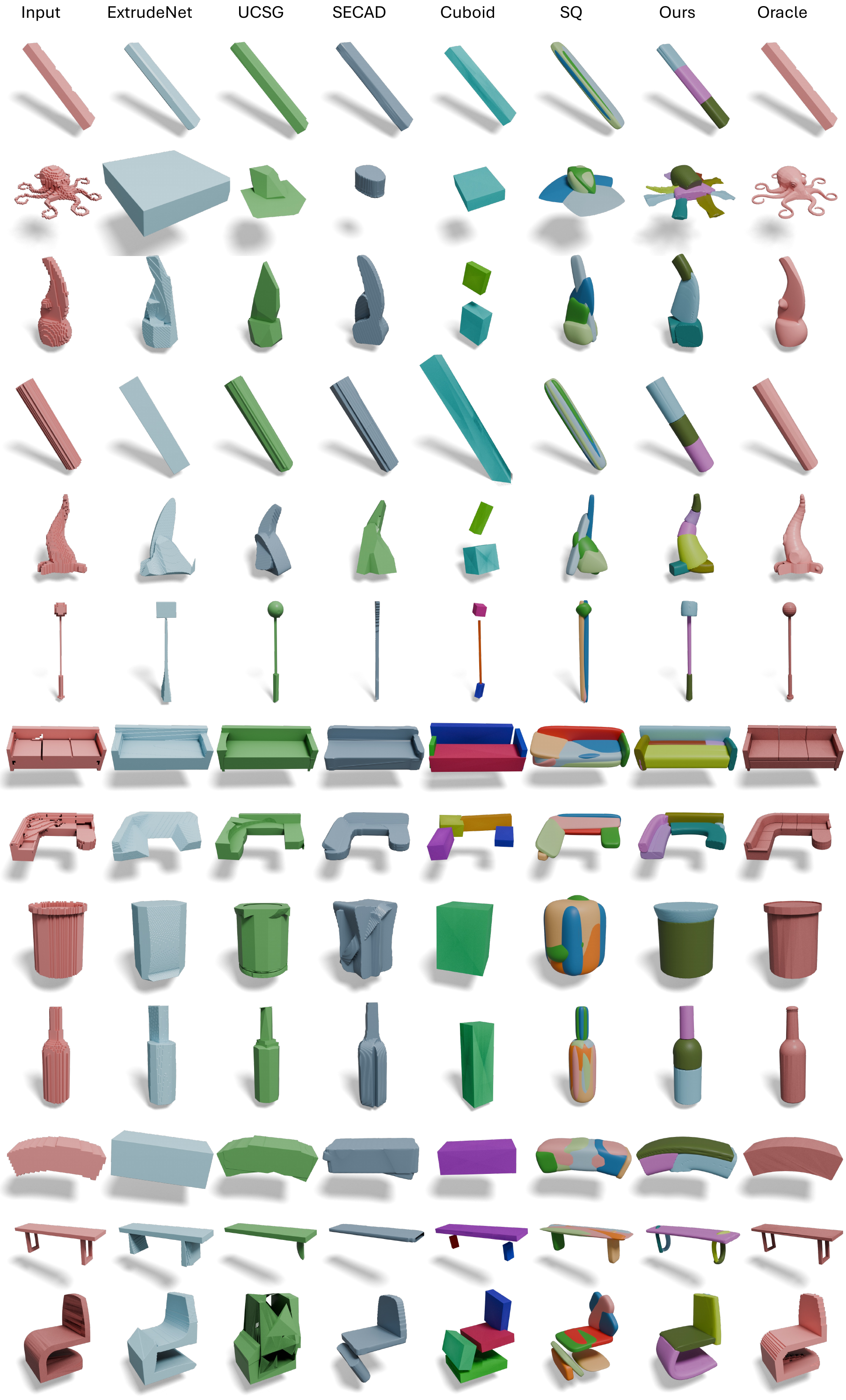}
    \caption{Additional qualitative results on Thingi10K~\cite{zhou2016thingi10k} and ShaepNet datasets~\cite{chang2015shapenet}. Our method provides reasonable abstractions on objects lack of sweep elements.}
    \label{fig:sup-sn-thingi10k}
\end{figure}

\end{document}